\documentclass[letterpaper, 10 pt, conference]{ieeeconf}

\IEEEoverridecommandlockouts    

\usepackage{graphics}           
\usepackage{epsfig}             
\usepackage{mathptmx}           
\usepackage{times}              
\usepackage{amsmath}            
\usepackage{amssymb}            
\usepackage{xspace}             
\usepackage[hang,flushmargin]{footmisc}
\usepackage{microtype}

\makeatletter                   
\let\NAT@parse\undefined        
\makeatother                    

\newcommand{\net}{AmodalFlowNet\xspace}
\newcommand{\metric}{Amodal Flow Quality\xspace}

\newcommand{\secref}[1]{Sec.~\ref{#1}}
\newcommand{\figref}[1]{Fig.~\ref{#1}}
\newcommand{\tabref}[1]{Tab.~\ref{#1}}

\usepackage[dvipsnames]{xcolor}
\definecolor{dark-green}{RGB}{12,80,12}

\newcommand{\custompar}[1]{%
    {\parskip=3pt\par\noindent\textit{#1}:\ }\ignorespaces%
}

\usepackage{url}
\usepackage{dsfont}
\usepackage{amsmath}
\usepackage{amssymb}
\usepackage{cuted}
\usepackage{multirow} 
\usepackage{multicol}
\usepackage{flushend}

\let\labelindent\relax
\usepackage{enumitem}
\usepackage{tabularx}

\usepackage{booktabs}
\usepackage{xspace}
\usepackage{caption}
\usepackage{graphicx} 
\usepackage{color}
\usepackage{siunitx}
\usepackage{cite}
\usepackage[pdfencoding=auto, hidelinks]{hyperref} 

\iffalse 
  \newcommand{\todo}[1]{\noindent}
\else
  \newcommand{\todo}[1]{\textcolor{red}{\bf [Todo: #1]}}
  
\fi

\title{\LARGE \bf Amodal Optical Flow}

\author{Maximilian Luz$^{*,1}$,
        Rohit Mohan$^{*,1}$,
        Ahmed Rida Sekkat$^{2}$,\\
        Oliver Sawade$^{2}$,
        Elmar Matthes$^{2}$,
        Thomas Brox$^{1}$,
        and~Abhinav Valada$^1$
\thanks{$^*$These authors contributed equally.}%
\thanks{$^1$Department of Computer Science, University of Freiburg, Germany}
\thanks{$^2$IAV GmbH, Germany}%
\thanks{This work was funded by the German Research Foundation Emmy Noether Program grant number 468878300 and an academic grant from NVIDIA.}}

\begin{document}

\maketitle
\thispagestyle{empty}
\pagestyle{empty}

\begin{abstract}
%
Optical flow estimation is very challenging in situations with transparent or occluded objects. In this work, we address these challenges at the task level by introducing Amodal Optical Flow, which integrates optical flow with amodal perception.
Instead of only representing the visible regions, we define amodal optical flow as a multi-layered pixel-level motion field that encompasses both visible and occluded regions of the scene.
To facilitate research on this new task, we extend the AmodalSynthDrive dataset to include pixel-level labels for amodal optical flow estimation. We present several strong baselines, along with the \metric metric to quantify the performance in an interpretable manner.    Furthermore, we propose the novel \net as an initial step toward addressing this task. \net consists of a transformer-based cost-volume encoder paired with a recurrent transformer decoder which facilitates recurrent hierarchical feature propagation and amodal semantic grounding.
We demonstrate the tractability of amodal optical flow in extensive experiments and show its utility for downstream tasks such as panoptic tracking. We make the dataset, code, and trained models publicly available at \url{http://amodal-flow.cs.uni-freiburg.de}.
\end{abstract}

\section{Introduction}


Optical flow estimates the apparent motion patterns of objects between two consecutive images of a sequence.
This fundamental vision problem has widespread applications, including localization~\cite{cattaneo2020cmrnet++}, autonomous driving~\cite{gosala2023skyeye}, and object tracking~\cite{buchner20223d}.
Over the years, several groundbreaking methods have been proposed~\cite{hornDeterminingOpticalFlow1981, broxHighAccuracyOptical2004, dosovitskiyFlowNetLearningOptical2015, ilgFlowNetEvolutionOptical2017, sunPWCNetCNNsOptical2018, teedRAFTRecurrentAllPairs2020, jiangLearningEstimateHidden2021, xuGMFlowLearningOptical2022, shiFlowFormerMaskedCost2023}. One remaining challenge is still \textit{transparency} (e.g., seeing through windows). Optical flow, by itself, cannot represent transparency accurately as it can only provide a single displacement vector per pixel.
Another remaining challenge is \textit{occlusion} and expressing the occlusion relationship of objects adeptly, which leads to incomplete and inaccurate motion patterns, particularly at occlusion boundaries.
To address these challenges and enhance motion analysis, we exploit amodal perception~\cite{zhuSemanticAmodalSegmentation2017, li2016amodal, mohan2022amodal}, which aims to perceive objects as a whole, even when parts of them are occluded.

\begin{figure}
   \centering
    \includegraphics[width=0.9\linewidth]{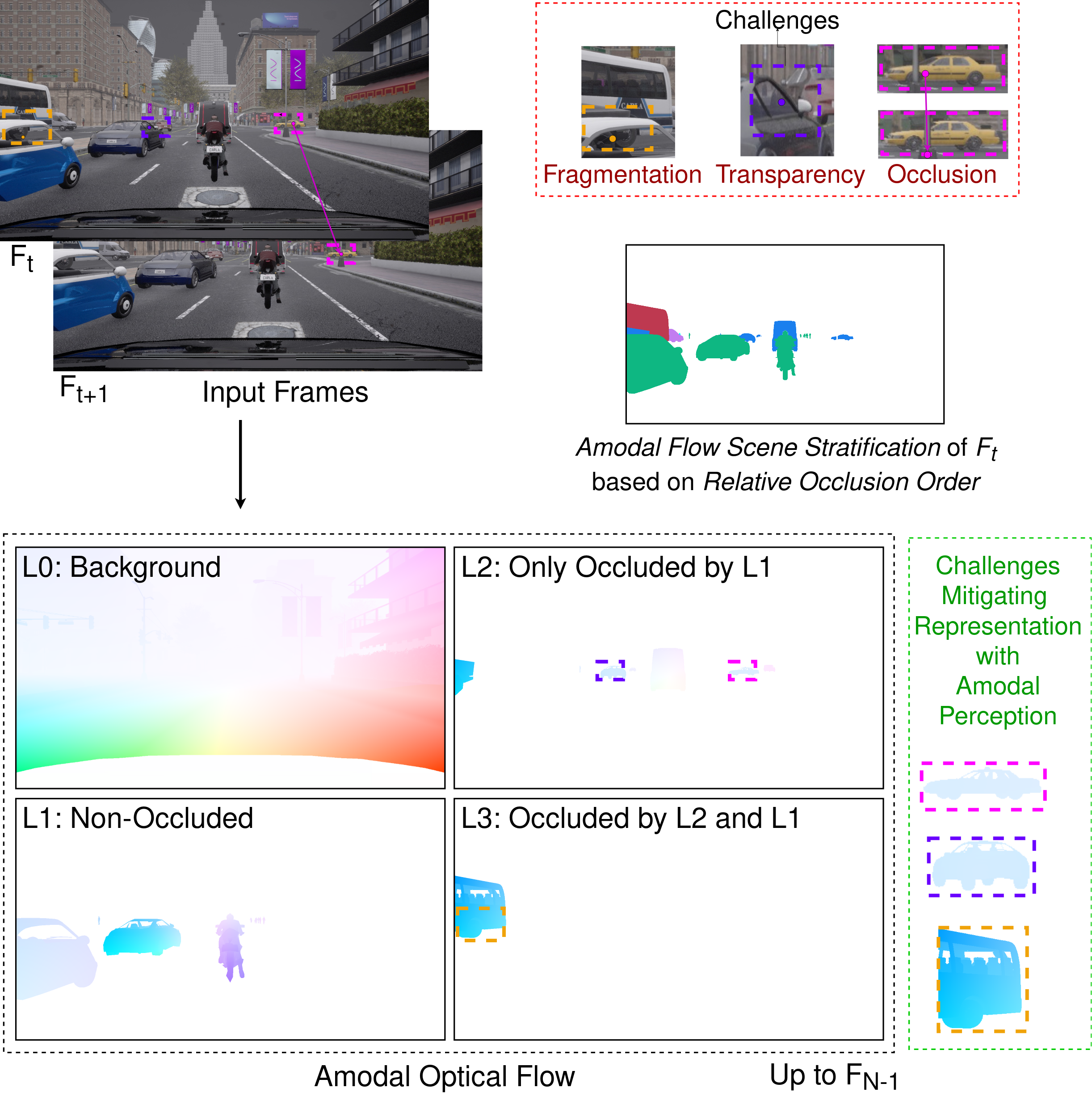}
    \caption{Illustration of \textit{Amodal Optical Flow}, which aims to predict a multi-layered pixel-level motion field encompassing both visible and occluded regions of the scene. This task can represent transparent and partially occluded objects while also reducing the fragmentation of object segments through amodal (visible + occluded) motion representation of scene elements.} 
    \label{fig:paper-teaser}
    \vspace{-0.5cm}
\end{figure}

In this work, we introduce amodal optical flow that seamlessly incorporates the principles of optical flow with amodal perception. As illustrated in \figref{fig:paper-teaser}, amodal optical flow aims to predict a set of motion fields at the pixel level, where each pixel is distinctly associated with both the visible and occluded regions of different scene elements across consecutive frames.
This pixel association is defined through the use of amodal masks of the scene elements (refer to \secref{subsec:taskdef}).
Fundamentally, amodal optical flow requires the delineation of continuous object boundaries that encompass both visible and occluded regions.
Thereby, it directly addresses the fragmentation problem encountered by most optical flow estimation methods, where refinement follows texture boundaries and edges rather than the true object boundaries (cf.~\cite{zhouSAMFlowEliminatingAny2023}).

By encouraging explicit reasoning about occlusions, amodal optical flow facilitates a more comprehensive handling thereof and lends itself more naturally to multi-frame sequence predictions. Furthermore, it moves away from the single-layer prediction that is typical in optical flow, adopting multi-layer outputs that are better suited to handle the complex interactions and overlaps between scene elements, including transparency. This is particularly relevant for scene understanding in robotics. Explicitly modeling the motion of individual scene elements is closer to the 3D nature of the scene, thus allowing for more informed, accurate decisions.  Object tracking is a concrete example that can benefit from amodal optical flow.

Our main contributions are twofold. First, we formulate the amodal optical flow estimation task, and to facilitate research, we provide a labeled dataset, an appropriate evaluation metric, and baselines. Given the inherent complexities of amodal perception and the challenges associated with acquiring real-world ground truth labels, we propose a synthetic dataset, which offers the advantage of controlled scenarios and perfect precision of annotations. We extend the recently introduced AmodalSynthDrive dataset~\cite{sekkat2023} with amodal optical flow ground truth, enabling other tasks to leverage amodal flow to improve performance and vice versa. Second, we propose the novel \net architecture, which predicts amodal flow in a recurrent and occlusion-ordered manner. Notably, it keeps the number of objects or occlusions in a scene flexible. It separately predicts both the visible and invisible regions of objects (i.e., amodal masks), as well as amodal semantic labels. We show the tractability and effectiveness of our approach in several experiments. Furthermore, we compare amodal optical flow and traditional optical flow in terms of their efficacy for panoptic tracking. The result demonstrates that the tracking approach based on amodal optical flow compares favorably. 
We will make the code and trained models publicly available at \mbox{\url{http://amodal-flow.cs.uni-freiburg.de}}.

\section{Related Work}

\custompar{Amodal Perception}
Amodal perception is an integral part of human cognitive abilities~\cite{smith2013cognitive}.
Consequently, various computer vision tasks have been proposed, often as counterparts to classical modal problems. Amodal bounding box prediction~\cite{dengAmodalDetection3D2017} and amodal instance segmentation~\cite{li2016amodal} aim to infer the true extent of objects in a scene, including any occluded parts.
Zhu~\emph{et~al.}~\cite{zhuSemanticAmodalSegmentation2017} extend the latter with semantic information; however, they still only consider salient regions of the image.
Contrastively, Purkait~\emph{et~al.}~\cite{purkaitSeeingThingsExtending2019} adapt semantic segmentation to the amodal setting more directly, performing a standard modal segmentation of the background assuming no foreground objects are present and only handling the latter amodally.
Amodal panoptic segmentation~\cite{mohan2022amodal,mohan2022perceiving} reintroduces instance information to this while keeping the separation of foreground and background classes, creating a holistic amodal segmentation approach.
Due to the inherent complexity of amodal perception, however, only a few approaches go beyond the prediction of object shapes and masks, largely focusing on appearance (e.g., \cite{ehsaniSeGANSegmentingGenerating2018,yanVisualizingInvisibleOccluded2019}).
Notably, Dhamo~\emph{et~al.}~\cite{dhamoPeekingObjectsLayered2019} consider the full scene and additionally estimate per-object depth.
We refer to the survey by Ao~\emph{et~al.}~\cite{aoImageAmodalCompletion2023} for further discussion on amodal perception.
%
%

\custompar{Optical Flow Estimation}
While nowadays vastly surpassed by neural networks, classical optical flow estimation approaches (e.g., \cite{blackRobustDynamicMotion1991,broxHighAccuracyOptical2004,sunSecretsOpticalFlow2010,sunQuantitativeAnalysisCurrent2014}) still contain invaluable knowledge.
Particularly relevant to amodal optical flow are techniques that decompose the scene into multiple overlapping layers, each with its own flow field and mask.
Although earlier methods (e.g., \cite{wangRepresentingMovingImages1994,jojicLearningFlexibleSprites2001,kannanGenerativeModelDense2006,sunLayeredImageMotion2010}) show limited amodal reasoning through appearance reconstruction in occluded regions, later methods (e.g., \cite{sunLayeredSegmentationOptical2012,sunFullyConnectedLayeredModel2013,sevilla-laraOpticalFlowSemantic2016}) solely aim to improve their modal capabilities.
Rather than constructing an amodal representation of optical flow, a major motivation for their development was the composition of complex motion via multiple simpler models. Consequently, little evaluation has been performed targeting amodal aspects, and, to our knowledge, no amodal optical flow metrics or evaluation datasets exist.
The introduction of neural network-based flow estimation methods has made explicit motion decomposition as a modeling tool less important.
Instead, recent methods have focused on recurrent refinement~\cite{teedRAFTRecurrentAllPairs2020}, motion aggregation for ambiguous or otherwise difficult matches~\cite{jiangLearningEstimateHidden2021}, better features for correspondence estimation~\cite{xuGMFlowLearningOptical2022,xuUnifyingFlowStereo2023}, and improved matching cost representations~\cite{shiFlowFormerMaskedCost2023}.\looseness=-1

\custompar{Combining Motion and Segmentation}
Even though we attempt to retain the more fundamental nature of optical flow by disentangling it from detailed object knowledge, some object reasoning is nevertheless required to address the amodal aspects.
Therein, amodal optical flow is related to the detection and segmentation of independently moving objects from image sequences \cite{broxObjectSegmentationLong2010,ochsSegmentationMovingObjects2014,fragkiadakiLearningSegmentMoving2015}.
While amodal approaches have been proposed in this regard (e.g., \cite{lamdouarSegmentingInvisibleMoving2021,xieSegmentingMovingObjects2022}), optical flow approaches so far only employ segmentation for direct modal flow improvements (e.g., \cite{tsaiVideoSegmentationObject2016,chengSegFlowJointLearning2017,zhouSAMFlowEliminatingAny2023}).

\section{Amodal Optical Flow}
\label{subsec:taskdef}

The objective of amodal optical flow estimation is to compute motion patterns for scene components within both visible and occluded regions across consecutive frames, \( I_t \) and \( I_{t+1} \). This involves determining a set of motion fields, \( \mathcal{F} \), where each \( F_i \) in \( \mathcal{F} \) corresponds to a visible scene component or a potentially occluded scene element. For each pixel \( p \) in \( I_t \), its probable positions in \( I_{t+1} \) are given by \( p + F_i(p) \), where \( F_i \) represents one of the possible displacement vectors \( (u, v) \) from the set. To accommodate the complexity of potentially diverse motion field predictions for a single pixel, considering both visible and occluded regions of scene elements, we employ the notion of relative occlusion order~\cite{mohan2022perceiving}. This facilitates the development of a scene stratification strategy for amodal flow that ensures a consistent order in the motion field set. This framework organizes scene elements hierarchically based on their visibility and occlusion potential.


\custompar{Amodal Flow Scene Stratification}
We introduce two fundamental categories of scene elements, inspired by panoptic segmentation: 
\begin{itemize}[topsep=0pt]
    \item \textit{Background:}  This category includes amorphous regions such as roads, walls, and static objects such as traffic signs and lights. 
    \item \textit{Traffic participants:} This category comprises movable objects such as pedestrians and cars. 
\end{itemize}
We further simplify the complexity of scenes by disregarding occlusions occurring solely within background elements. The primary focus of the task considers the interplay of occlusion between the background and traffic participants, as well as occlusions among different instances of traffic participants. This simplification ensures clarity and precision in predicting the motion fields. Based on these premises, we establish a layered stratification of the scene in frame \( I_t \) that consists of \( N \) levels, each representing a distinct occlusion degree:
\begin{itemize}[topsep=0pt]
    \item Level 0: Corresponds to the background.
    \item Level 1: Encompasses non-occluded traffic participants.
    \item Level 2 to Level \(N-1\): Each subsequent level includes objects occluded by at least one object from the preceding level and any objects from earlier levels. Notably, objects within the same level do not occlude each other.
\end{itemize}

This stratified structure allows a pixel to have multiple associated motion field vectors, each tied to a specific occlusion level. It is important to highlight that this strategy remains independent of the specific scene element classes, enabling the task to prioritize motion estimation without heavy reliance on object recognition. Moreover, this simplification encourages the potential for unsupervised approaches and adaptability across diverse environments, facilitating the emergence of innovative methods beyond our approach detailed in~\secref{sec:approach}, which capitalizes on the semantic understanding of objects. We set \( N \) at a maximum of 8, corresponding to the highest value observed in our dataset. \figref{fig:paper-teaser} shows a visual representation of this stratification strategy with an example scene along with the corresponding amodal optical flow ground truth for each occlusion order level.

\begin{figure*}
    \centering
    \includegraphics[width=\linewidth]{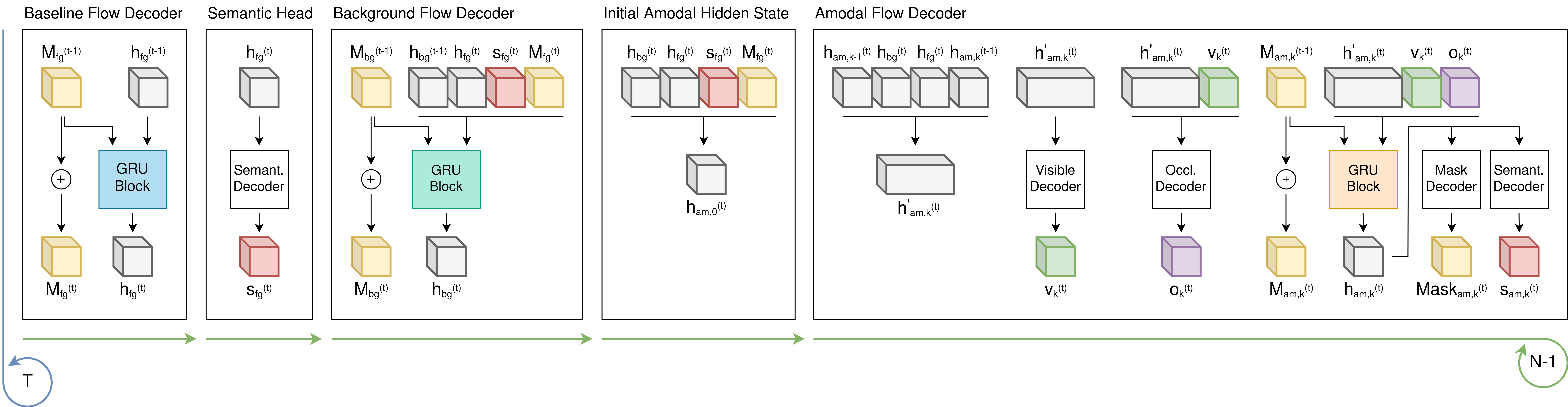}
    \caption{AmodalFlowNet architecture. Flow and corresponding amodal masks (yellow blocks) are estimated recurrently over both refinement steps (outer, blue arrow) and amodal layers (inner, green arrows). The decoder structure for the standard optical flow is retained from the baseline model. Additional semantic and mask predictions (red, green, and purple blocks) guide the network.}
    \label{fig:amodalflownet}
    \vspace{-0.4cm}
\end{figure*}

\custompar{Task Definition}
\label{subsec:taskdefinition}
For the given inputs \( I_t \) and \( I_{t+1} \), amodal optical flow aims to predict a series of motion field maps \(\{M_0, M_1, \ldots, M_{N-1}\}\), each corresponding to a specific occlusion level \(n\) within the scene. Every motion field map \(M_n\) has the same dimensions as the input, where each pixel value is represented by a triplet \(\langle i, u, v \rangle\). Here, \(i\) indicates whether the pixel is associated with a scene element segment (taking a value of 0 or 1), while \(u\) and \(v\) represent the components of the displacement vector at that point.
Formally, the prediction step can therefore be represented as
\begin{equation}
    f(I_t, I_{t+1}) \to \{M_0, M_1, \ldots, M_{N-1}\}.
\end{equation}
\custompar{Evaluation Metric}
\label{subsec:evalmetric}
To evaluate amodal optical flow estimation, we introduce a unified metric called Amodal Flow Quality (AFQ). This metric jointly assesses both the amodal motion field prediction and the class-agnostic segmentation quality of scene elements in an interpretable manner. This is achieved by combining the weighted area under the curve (WAUC)~\cite{richter2017playing} metric utilized in the evaluation of optical flow with the mean intersection over union (mIoU) metric~\cite{hurtado2022semantic} which is commonly used for measuring segmentation quality. We denote the predicted amodal optical flow for occlusion level $n$ as $M_n^p$ and the ground truth amodal optical flow for occlusion level $n$ as $M_n^g$. We then define the WAUC for each occlusion level $n$ that integrates over distance thresholds between 0 and 5 pixels as follows:
\begin{align}
    \mathrm{WAUC}_n &= \frac{\sum_{i=1}^{100} w_i \sum_{j} [e_j \leq \delta_i]}{C \cdot \sum_{i=1}^{100} w_i},\\
e_j &= \left\| \vec{v}_{M_n^p}(j) - \vec{v}_{M_n^g}(j) \right\|_2,
\end{align}
where $e_j$ is the end point error at pixel j, $\vec{v}_{.}(j)$ represent the motion vectors containing both the u and v components and $\delta_i = \frac{i}{20}$. $[.]$ is the Iverson bracket, $C$ is the total number of pixels, and $w_i = 1 - \frac{i-1}{100}$.  
Similarly, we define IoU for each occlusion level $n$ as follows:
\begin{equation}
    \mathrm{IoU}_n = \frac{\mathrm{TP}}{\mathrm{TP} + \mathrm{FP} + \mathrm{FN}},
\end{equation}
where TP, FP, and FN are the number of pixels correctly segmented, falsely segmented, and missed in the segmentation, respectively. Having computed the individual WAUC and IoU values for each occlusion level, we compute AFQ as the geometric mean of the mean IoU (mIoU) and the mean WAUC (mWAUC), both averaged over all occlusion levels,
\begin{equation}
    \mathrm{AFQ} = \sqrt{\mathrm{mWAUC} \cdot \mathrm{mIoU}}.
\end{equation}
Hereby, mIoU and mWAUC are defined as
\begin{align}
    \mathrm{mIoU} &= \frac{1}{\sum_{n=1}^{N-1} w_n} \sum_{n=1}^{N-1} w_n \mathrm{IoU}_n, \\
    \mathrm{mWAUC} &= \frac{1}{\sum_{n=0}^{N-1} w_n} \sum_{n=0}^{N-1} w_n \mathrm{WAUC}_n,
\end{align}
with exponentially decaying weights
\begin{equation}
    w_n = \exp \left(- \max\left(-\frac{n - k}{N - 1 - k} \log(w_{N-1}),\ 0\right) \right),
\end{equation}
where $N$ is the total number of occlusion levels, $k=3$ the number of amodal levels with equal weighting, and $w_{N-1}=0.25$ the weight of the last layer. The choice of $k$ is determined by the dataset's relative occlusion order distribution.

\section{Dataset}
\label{sec:dataset}

AmodalSynthDrive~\cite{sekkat2023} is the first comprehensive dataset for multi-task multi-modal amodal perception tailored to the automotive domain. It consists of \num{60000} multi-view camera images, 3D bounding boxes, LiDAR data, odometry, and amodal semantic/instance/panoptic segmentation annotations, distributed across \num{150} distinct driving sequences, with over 1 million object annotations. It covers various scenarios involving diverse traffic, weather conditions, and lighting conditions. In this work, we extend this dataset with amodal optical flow ground truth for images of the original training, validation, and test splits, consisting of \num{105} video sequences with \num{42000} images, \num{15} video sequences with \num{6000} images, and \num{30} video sequences with \num{12000} images, respectively.

\begin{figure}
    \centering
    \includegraphics[width=\columnwidth]{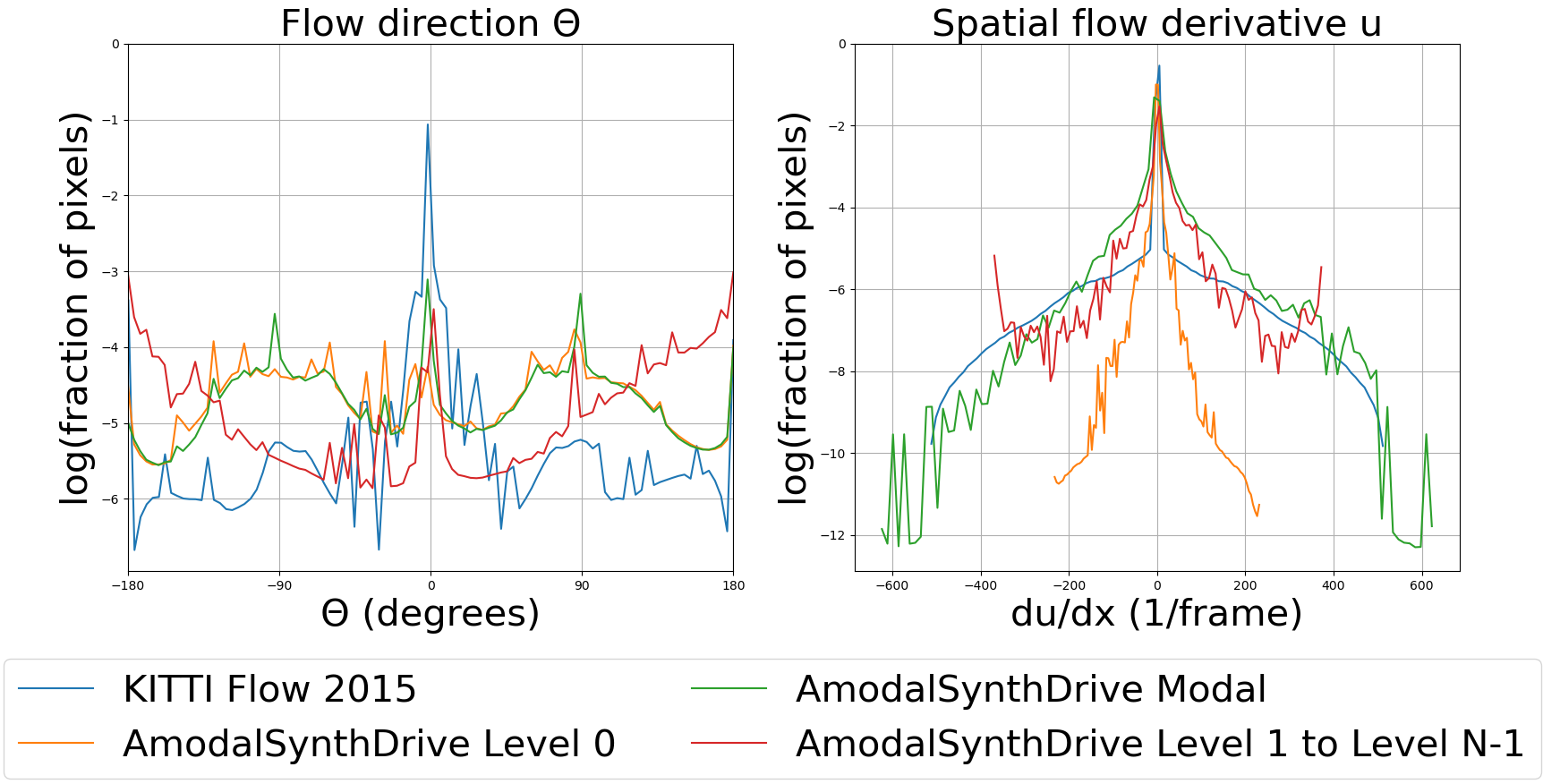}
    \caption{Log histograms of the optical flow direction and the spatial derivative of the horizontal optical flow velocity $u$. The modal statistics of AmodalSynthDrive (green) demonstrate similarities to KITTI (blue), indicating similar motion patterns, albeit the existence of spatial irregularity. This distinction can be attributed to the fact that the optical flow ground truth in our dataset is characterized by its high level of detail and precision.}
    \label{fig:flow_statistics}
    \vspace{-0.3cm}
\end{figure}

We compute the amodal optical flow by individually rendering all visible objects, excluding the background, frame by frame. In the rendering procedure, we individually extract ground truth labels for each object. This ground truth encompassed the data required to compute the optical flow for each individual object separately, namely depth maps, instance segmentation, and precise positions and rotations.
For every object in the scene, we compute both its displacement between two consecutive frames and the corresponding camera displacement representing the displacement transformations. We also compute the amodal point cloud 
by utilizing the corresponding amodal depth maps. 
By applying the displacement transformations, we generate the amodal point cloud that underwent the aforementioned movements. This allows us to derive 3D movement vectors for each pixel representing the object in the amodal instance segmentation. These 3D vectors are then reprojected into the different image planes to obtain modal and amodal optical flow. 
To gain a more comprehensive understanding of the optical flow ground truth in our dataset, we perform statistical analyses as depicted in \figref{fig:flow_statistics} on both our dataset and the KITTI dataset~\cite{KITTI_Flow_2015}. Our motivation for this analysis is grounded in the idea that when the statistical characteristics of the computed flow align, it hints at a likely similarity in the depicted scenes and motions.

\section{Methodology}
\label{sec:approach}

In order to demonstrate the feasibility of amodal optical flow estimation, we present two baselines.
To this end, we adapt three pre-trained standard optical flow estimation methods to the amodal setting: GMA~\cite{jiangLearningEstimateHidden2021}, GMFlow+~\cite{xuGMFlowLearningOptical2022,xuUnifyingFlowStereo2023}, and FlowFormer++~\cite{shiFlowFormerMaskedCost2023}.
%
We chose these approaches specifically due to their respective contributions:
GMA represents the natural successor to the transformational RAFT~\cite{teedRAFTRecurrentAllPairs2020} approach, extending it with a motion aggregation module to aid refinement.
GMFlow+ provides convincing results for the standard optical flow task, resulting from its strong transformer-based feature extractor.
FlowFormer++ similarly achieves state-of-the-art performance but uses a more expressive matching cost representation by applying a transformer encoder-decoder network to the cost volume instead.
We presume the latter to be particularly valuable for the amodal optical flow task.
Our proposed \net module integrates well with all three approaches.

\custompar{\net Architecture}
Our core amodal optical flow prediction is structured in a recurrent hierarchical manner using three flow decoder modules as shown in \figref{fig:amodalflownet}.
In summary, we first estimate the standard optical flow using the respective baseline approach, representing the first decoder.
From this, the second decoder predicts the flow field for the background ($M_0$).
Lastly, the third decoder recurrently predicts amodal flow and masks for all amodal object layers, ordered from front to back (i.e., $M_1, M_2, \ldots, M_{N-1}$), again employing the previously predicted standard flow field as an additional input.
This recurrent hierarchical structure allows the third decoder to carry over information from the previous layers, enabling occlusion-aware prediction.

Notably, both baselines employ a RAFT-like recurrent refinement framework~\cite{teedRAFTRecurrentAllPairs2020}.
This signifies that flow is estimated over the course of multiple refinement iterations, each predicting a residual flow update.
The full flow output is then computed as the sum over all residual updates.
In this context, the aforementioned amodal estimation process takes place during each (outer) refinement iteration.
Furthermore, we exploit the hidden state of the recurrent update module proposed by RAFT to exchange information between the different decoders.
In particular, we maintain $N + 1$ hidden states: one for the standard flow ($\smash{h_{\text{fg}}^t}$), one for the background ($\smash{h_{\text{bg}}^t}$), and one for each amodal object layer ($\smash{h_{\text{am},k}^t},\ k \in \{1,\ldots,N-1\}$).
While $\smash{h_{\text{fg}}^t}$ is kept as-is from the baseline, the hidden state of the background decoder $\smash{h_{\text{bg}}^t}$ is fused with the hidden state of the standard flow decoder $\smash{h_{\text{fg}}^t}$ before being passed into the background update module in each outer refinement iteration.
Similarly, the hidden state of each amodal layer $\smash{h_{\text{am},k}^t}$ is fused with the hidden states corresponding to the standard flow $\smash{h_{\text{fg}}^t}$, background flow $\smash{h_{\text{bg}}^t}$, and previous layer $\smash{h_{\text{am},k-1}^t}$.
We construct $\smash{h_{\text{am},0}^t}$ as a fusion of hidden states from the standard ($\smash{h_{\text{fg}}^t}$) and background ($\smash{h_{\text{bg}}^t}$) decoders.
This therefore creates a network that is recurrent over both time steps and amodal layers.

To provide additional guidance to our architecture, we predict several auxiliary targets and consider them in our training loss.
For the object layers ($1$ to $N$), we decompose the amodal masks into visible and occluded regions akin to \cite{mohan2022amodal,tran2022aisformer}, predicting both separately from the fused hidden state of the layer.
The decomposed masks are then fed back to the update module to aid in the estimation of the complete amodal mask and flow.
Additionally, we predict semantic labels for both the full frame and the amodal object layers.
While the semantic labels of the amodal layers are only used as an additional training target, the semantic labels of the full frame are fed to the update module of the background flow and the initial object layer hidden state $\smash{h_{a,0}^t}$.

\custompar{Training Loss}
We adapt the sequence loss introduced by RAFT~\cite{teedRAFTRecurrentAllPairs2020} by adding amodal and background flow, semantic, and mask terms to the per-iteration loss.
In particular, we employ a $L_1$ end-point loss term for each flow output per iteration, forcing amodal flow predictions to zero outside the area of amodal objects.
The semantic labels are trained with a cross-entropy loss, and the visible and occlusion masks with a binary cross-entropy loss.
Therefore, the total loss can be written as follows:
\begin{equation}
    \mathcal{L} =
        \sum_{t=0}^{T} \gamma^{T - t}
        \left(
            F_{t, \text{fg}}
            + \mathcal{L}_{\text{am},t}
        \right),
\end{equation}
\begin{equation}
    \mathcal{L}_{\text{am},t} =
        F_{t, \text{bg}}
        + S_{t, \text{fg}}
        + \sum_{k=1}^{N-1}
            F_{t, k} + S_{t, k}
            + M_{t, k, \text{vis}}
            + M_{t, k, \text{occ}},
\end{equation}
where $S_{t, \text{fg}}$ and $S_{t, k}$ are the cross-entropy losses for the full and $k$-th layer amodal segmentation labels, respectively; $M_{t, k, \text{vis}}$ and $M_{t, k, \text{occ}}$ the binary cross-entropy losses for the amodal visible and occlusion masks; and $F_{t, \text{fg}}$, $F_{t, \text{bg}}$, and $F_{t, k}$ the $L_1$ end-point error losses for the standard, background, and zero-extended amodal optical flow.

\section{Experimental Evaluation}
In this section, we first present the benchmarking results on the AmodalSynthDrive dataset~\cite{sekkat2023} in \secref{sec:benchmarking}, followed by an ablation study on the different architectural design choices of \net in \secref{sec:ablation}. We then evaluate the utility of amodal optical flow for the downstream task of panoptic tracking~\cite{hurtado2020mopt}. Finally, we present qualitative comparisons in \secref{sec:qualitative}.\looseness=-1

We trained all models for \num{120000} iterations with a batch size of six and four recurrent refinement steps. We initialize the modal base architecture of the baselines with pre-trained Sintel~\cite{Sintel} checkpoints, using Xavier initialization for the rest of the network.
For the remaining hyperparameters, we follow the Sintel fine-tuning strategy of FlowFormer++~\cite{shiFlowFormerMaskedCost2023}.


\subsection{Benchmarking Results}
\label{sec:benchmarking}

We present results from evaluations of the baselines on the validation and test sets of AmodalSynthDrive in \tabref{tab:baselines}, again using 4 recurrent iterations.
More specifically, we compare our FlowFormer++-based \net architecture (as described in \secref{sec:approach}) against GMA- and GMFlow+-based learned baselines without mask and semantic guidance (cf. M1 in \secref{sec:ablation} and \tabref{tab:ablarch}).
Additionally, we evaluate two strategies for performing simple non-learned flow infilling of occluded areas based on the amodal mask predictions from our network, one extending the border between visible and occluded regions to the occluded region, whereas the other uses the mean of the visible area.

Comparing learned and non-learned baselines, we observe that there is clearly a need for techniques with a more comprehensive understanding of objects and their motion to effectively address the amodal optical flow task.
In particular, AmodalGMFlow+ outperforms both non-learned baselines for pure optical flow (mWAUC) by over \SI{64}{\percent} and \SI{46}{\percent} (\SI{16.2}{pp.} and \SI{11.0}{pp.}) on the validation and test split, respectively, leading to clear gains in the AFQ score.
Notably, however, it performs worse than our GMA-based baseline, indicating that global reasoning (in this case through motion aggregation) is better suited to the amodal task than pure improvements in the feature representation.
Our complete \net achieves an additional \SI{20}{\percent} (\SI{6.7}{pp.}) improvement in AFQ over AmodalGMA on the test split.
%

\begin{table}
\setlength\tabcolsep{1.8pt}
\centering
\caption{Comparison of amodal optical flow performance on the AmodalSynthDrive dataset. All scores are in [\%].}
\label{tab:baselines}
\footnotesize
\begin{tabular}
{l|ccc|ccc}
\toprule
\multirow{3}{*}{Method} & \multicolumn{3}{c|}{Val Set} & \multicolumn{3}{c}{Test Set} \\ 
\cmidrule{2-7}
&  AFQ  & mWAUC & mIoU &  AFQ  & mWAUC & mIoU\\
\toprule
Near boundary & 21.7 & 20.1 & 23.5 & 20.1 & 18.8 & 21.5 \\
Mean & 24.3 & 25.3 & 23.5 & 22.5 & 23.6 & 21.5 \\
AmodalGMFlow+ & 31.2 & 41.5 & 23.5 & 27.3 & 34.6 & 21.5 \\
AmodalGMA & 41.6 & 43.7 & 39.6 & 32.9 & 34.8 & 32.8 \\
\net (ours) & \textbf{45.8} & \textbf{49.4} & \textbf{42.4} & \textbf{39.6} & \textbf{42.0} & \textbf{37.3} \\
\bottomrule
\end{tabular}
\end{table}

\subsection{Ablation Study}
\label{sec:ablation}

\begin{figure*}
    \centering
    \includegraphics[width=\linewidth]{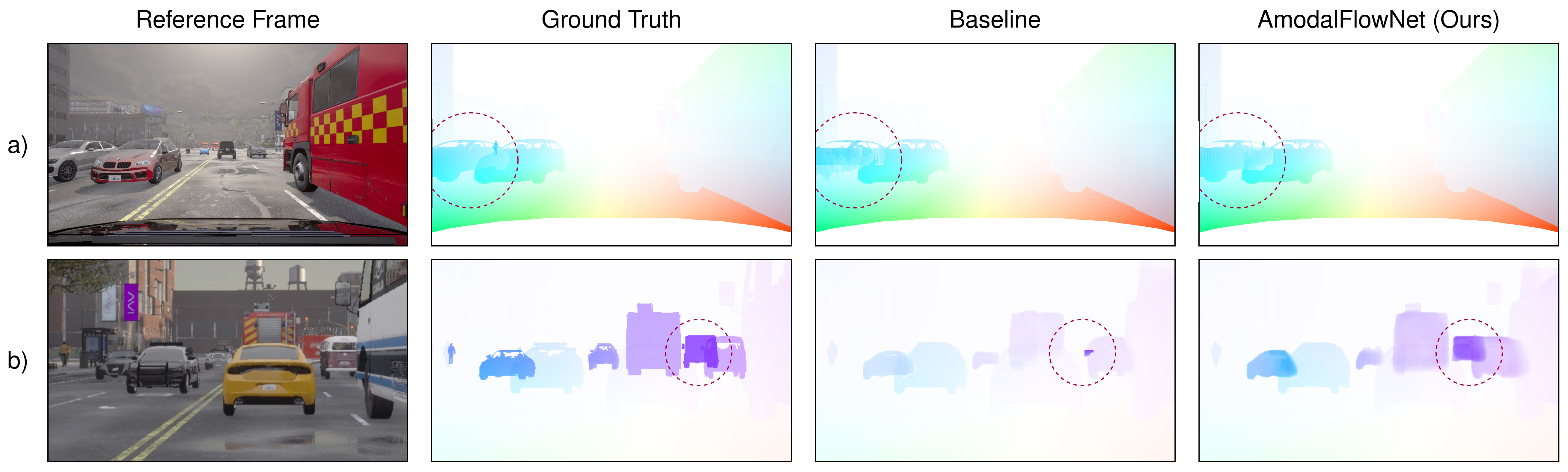}
    
    \caption{Qualitative comparison of amodal optical flow prediction from our proposed AmodalFlowNet with the baseline M1 on the AmodalSynthDrive dataset. For visualization, we sequentially superimpose the multi-layer amodal optical flow predictions, in order of $M_0, M_1, \ldots, M_{N-1}$}
    \label{fig:qual}
\vspace{-0.3cm}
\end{figure*}

We ablate the architecture of our proposed \net on the AmodalSynthDrive validation split to study the impact of our semantics- and mask-based guidance strategy. Results from this experiment are shown in \tabref{tab:ablarch}. We compare our model with full guidance, as described in \secref{sec:approach} (M3), against a simplified version (M2) without semantic training. Therein, the amodal semantic decoder is dropped completely, whereas the semantic decoder for the full scene is replaced by a decoder predicting a motion mask, representing the union over all amodal masks of any moving objects in the scene. Lastly, we also evaluate a model (M1) without any guidance, removing the prediction of semantics/motion masks as well as the decomposed amodal visible and occlusion masks entirely.

\begin{table}
\footnotesize 
\centering
\caption{Ablation study on the various architectural components of our proposed \net on the validation set of the AmodalSynthDrive dataset. All scores are in [\%].}
\label{tab:ablarch}
\begin{tabular}{l|l|ccc}
\toprule
Model & Guidance & AFQ  & mWAUC & mIoU \\
\midrule
M1 & --- & 44.6 & 48.9 & 40.7 \\
M2 & masks & 43.8 & 46.8 & 41.0 \\
M3 & masks + semantics & \textbf{45.8} & \textbf{49.4} & \textbf{42.4} \\
\bottomrule
\end{tabular}
\vspace{-0.4cm}
\end{table}

The results show that the prediction of decomposed amodal masks in combination with guidance through semantic information leads to significant performance improvements.
Notable hereby is that while predicting masks without semantics improves the IoU of the amodal masks, it has a detrimental effect on the optical flow quality as measured by the mWAUC score.
Including semantics, however, improves both mask and flow predictions.
We argue that providing semantic information likely helps our network to better localize and refine flow in amodal occluded regions.
Per-layer statistics confirm that improvements are largely due to better performance on the amodal object layers.
This demonstrates that our proposed \net can successfully associate shape and motion information and exploit them for both amodal flow and mask refinement.



\subsection{Exploiting Amodal Flow for Panoptic Tracking}

In this section, we present experimental results, showing the benefits of amodal optical flow for the complex downstream task of panoptic tracking by building upon a mask propagation framework~\cite{weber2021step} that leverages the output of a panoptic segmentation network to extract object masks. This framework uses optical flow to warp each predicted object between consecutive frames, followed by IoU matching with the Hungarian algorithm to assign consistent tracking IDs.

To study the effectiveness of amodal optical flow for panoptic tracking, we developed two distinct tracking models: Modal-MaskProp (MMP) and Amodal-MaskProp (AMP).
We employ APSNet~\cite{mohan2022amodal} for generating panoptic segmentation predictions for both models to ensure a fair comparison.
MMP uses FlowFormer++ for optical flow estimation. Conversely, AMP employs our proposed AmodalFlowNet and adapts the mask propagation framework to accommodate amodal optical flow and amodal object masks. This model takes both amodal and modal object masks from APSNet as input in conjunction with amodal optical flow predictions.
To identify the correct motion field map for a given object from the $N$ amodal optical flow layers, we compute the overlap between the amodal mask of the object and the masks of each layer in the amodal optical flow, selecting the layer with the highest overlap.
We then use this layer to warp the amodal object mask and perform IoU matching to assign consistent tracking IDs.
We evaluate the performance using the standard Panoptic Tracking (PAT) metric~\cite{fong2022panoptic}.

Results are shown in \tabref{tab:tracking}.
We observe that AMP outperforms MMP by margins of \SI{4.7}{\percent} and \SI{4.4}{\percent} on the validation and test sets, respectively.
Given that both models use the same panoptic segmentation architecture, the Panoptic Quality (PQ) component of the PAT metric remains the same.
The Tracking Quality (TQ) component of the PAT metric, which assesses object association performance across frames, demonstrates improvements of \SI{8.8}{\percent} for both validation and test sets.
These results substantiate the utility of amodal optical flow for enhancing tracking quality when paired with an appropriate amodal segmentation approach.

\begin{table}
\setlength\tabcolsep{5.0pt}
\centering
\caption{Comparison of panoptic tracking performance on the AmodalSynthDrive dataset. All scores are in [\%].}
\label{tab:tracking}
\footnotesize
\begin{tabular}
{l|ccc|ccc}
\toprule
\multirow{3}{*}{Method} & \multicolumn{3}{c|}{Val Set} & \multicolumn{3}{c}{Test Set} \\ 
\cmidrule{2-7}
&  PAT  & TQ & PQ &  PAT  & TQ & PQ\\
\toprule
Modal-MaskProp  & 50.6& 46.2 & 56.1 & 49.0 & 44.3 & 54.8 \\
Amodal-MaskProp & \textbf{53.0} &  \textbf{50.3} &  \textbf{56.1} &  \textbf{51.2} &  \textbf{48.2} & \textbf{54.8} \\
\bottomrule
\end{tabular}
\vspace{-0.3cm}
\end{table}

\subsection{Qualitative Evaluation}
\label{sec:qualitative}


\figref{fig:qual} presents qualitative comparisons of the amodal optical flow prediction from the proposed AmodalFlowNet with the M1 baseline (see \secref{sec:ablation}). We observe that AmodalFlowNet distinguishes between visible and occluded scene elements, as highlighted by the red circle, owing to its amodal semantic grounding coupled with the recurrent hierarchical feature propagation from background to amodal flow layer $M_{N-1}$. However, it falls short of providing a complete amodal flow profile for the occluded objects, as seen in \figref{fig:qual}(b). Despite its limitations, we believe that it poses a promising baseline for future research on amodal optical flow. \looseness=-1
\section{Conclusion}

In this work, we introduced the novel amodal optical flow estimation task, bringing optical flow to the invisible and occluded regions of scenes and therewith extending it to the amodal setting.
To enable quantitative analysis and evaluation, we formulated the amodal flow quality metric.
We demonstrated the feasibility of the task by extending the AmodalSynthDrive dataset with ground truth labels for amodal optical flow and by providing the \net architecture and comparing it with other learned and non-learned baselines.
Ablations validate the choices made in our architecture, in particular, that guidance through semantic information and decomposed amodal masks provides valuable information and structure to the network.
Lastly, we show that the amodal optical flow estimated from our \net method is indeed beneficial to panoptic tracking as a downstream application.
We therefore believe that amodal optical flow shows great potential for the robotics community and will prove valuable for aiding dynamic scene understanding.

\bibliographystyle{IEEEtran}
\bibliography{root.bib}

\end{document}